\documentclass[letterpaper, 10 pt, conference]{ieeeconf}  

\IEEEoverridecommandlockouts                              

\usepackage[utf8]{inputenc}

\usepackage[hidelinks]{hyperref}
\usepackage[style=ieee,hyperref,natbib=true,backend=bibtex,firstinits,doi=false,%
     mincitenames=1,maxcitenames=2,maxbibnames=99,sorting=none,terseinits=false,hyperref=true]{biblatex}
\bibliography{references.bib}
\renewbibmacro*{bbx:savehash}{}
\defbibheading{bibliography}[\bibname]{\section*{References}}

\usepackage{balance}
\usepackage{url}
\usepackage{graphicx}
\usepackage[usenames,dvipsnames]{xcolor}
\usepackage[draft]{fixme}
\fxsetup{theme=color}

\usepackage{amsmath}
\usepackage{amssymb}
\usepackage{textcomp}
\usepackage{siunitx}

\usepackage{tabularx}
\usepackage{booktabs}
\usepackage{arydshln}
\usepackage{multirow}
\usepackage{bm}
\usepackage[capitalize]{cleveref}
\crefname{section}{Sec.}{Sections}

\usepackage[nolist]{acronym}
\begin{acronym}
\acro{ALS}{Aerial Laser Scanning}
\acro{APE}{absolute position error}
\acro{ATE}{absolute trajectory error}
\acro{dof}[DoF]{degrees of freedom}
\acro{ekf}[EKF]{Extended Kalman-Filter}
\acro{ICP}{Iterative Closest Point}
\acro{GICP}{Generalized \acs{ICP}}
\acro{GMM}{Gaussian Mixture Model}
\acro{GN}{Gauss-Newton}
\acro{GPS}{global positioning system}
\acro{gnss}[GNSS]{global navigation satellite system}
\acro{IMU}{inertial measurement unit}
\acro{iou}[IoU]{Intersection over Union}
\acro{LM}{Levenberg-Marquardt}
\acro{LIO}{LiDAR-Inertial-Odometry}
\acro{LiDAR}{Light Detection and Ranging}
\acro{MOM}{Mutually Orthogonal Metric}
\acro{NN}{nearest neighbor}
\acro{PGO}{Pose Graph Optimization}
\acro{RMS}{root-mean-squared}
\acro{RMSE}{\acs{RMS} error}
\acro{RANSAC}[RANSAC]{random sample consensus}
\acro{RGB}{red, green, blue}
\acro{SAR}{search and rescue}
\acro{SLAM}{Simultaneous Localization and Mapping}
\acro{surfel}{surface element}
\acro{TLS}{terrestrial \acs{LiDAR} scanner}
\acro{ToF}{time-of-flight}
\acro{UAV}[UAV]{unmanned aerial vehicle}
\acro{UGV}{unmanned ground vehicle}
\acro{ukf}[UKF]{Unscented Kalman-Filter}
\acro{VO}[VO]{visual odometry}
\acro{VIO}[VIO]{visual inertial odometry}
\acro{VSLAM}[vSLAM]{visual \acs{SLAM}}
\acro{voxel}{volume element}
\acro{lod}[LoD]{Level of Detail}
\acro{DEM}{digital elevation map}
\acro{cgml}[CityGML]{City Geography Markup Language}
\end{acronym}

\usepackage{tikz}
\usepackage{pgfkeys}
\usetikzlibrary{arrows}
\usetikzlibrary{arrows.meta}
\usetikzlibrary{decorations.pathreplacing, positioning, calc}
\usetikzlibrary{positioning,calc}
\usetikzlibrary{fit}
\usetikzlibrary{shapes.arrows}
\usetikzlibrary{shapes.geometric}

\newcommand{\abs}[1]{\lvert#1\rvert}

\newcommand{\wrt}{w.r.t.~}

\newcommand{\eg}{e.g.,\ }

\DeclareMathOperator*{\argmin}{arg\,min}

\newcolumntype{Z}{>{\centering\arraybackslash}X}
\newcolumntype{L}{>{\hspace{1pc}\raggedright\arraybackslash}X}
\newcolumntype{K}[1]{>{\hspace{1pc}\raggedright\arraybackslash}m{#1}}
\newcolumntype{N}{>{\raggedright\arraybackslash}X}
\newcolumntype{C}[1]{>{\centering\arraybackslash\hspace{0pt}}m{#1}}
\usepackage[export]{adjustbox}

\title{\LARGE \bf
LiDAR-based Registration against Georeferenced Models\\
for Globally Consistent Allocentric Maps
}

\author{Jan Quenzel~${}^{a,b}$, Linus T. Mallwitz~${}^{a}$, Benedikt T. Arnold~${}^{c}$, and Sven Behnke~${}^{a,b,d}$
\thanks{${}^a$~Autonomous Intelligent Systems Group, Computer Science Institute VI -- Intelligent Systems and Robotics -- and ${}^b$~Center for Robotics and Lamarr Institute for Machine Learning and Artificial Intelligence, University of Bonn, Germany;
${}^c$~Fraunhofer FIT, Germany;
${}^d$~Fraunhofer IAIS, Germany;
 {\tt\small quenzel@ais.uni-bonn.de}}%
}

\begin{document}

\maketitle
\thispagestyle{empty}
\pagestyle{empty}
\begin{tikzpicture}[remember picture,overlay]
  \node[anchor=north,align=center,font=\sffamily\small,yshift=-0.4cm] at (current page.north) {%
  \textbf{Accepted final version.} IEEE International Symposium on Safety, Security, and Rescue Robotics (SSRR), New York City, USA, November 2024
  };
\end{tikzpicture}%

\begin{abstract}
Modern \acp{UAV} are irreplaceable in \ac{SAR} missions to obtain a situational overview or provide closeups without endangering personnel.
However, \acp{UAV} heavily rely on \ac{gnss} for localization which works well in open spaces, but the precision drastically degrades in the vicinity of buildings.
These inaccuracies hinder aggregation of diverse data from multiple sources in a unified georeferenced frame for \ac{SAR} operators.

In contrast, \acs{cgml} models provide approximate building shapes with accurate georeferenced poses.
Besides, LiDAR works best in the vicinity of 3D structures.
Hence, we refine coarse \ac{gnss} measurements by registering LiDAR maps against \acs{cgml} and \ac{DEM} models as a prior for allocentric mapping.
An intuitive plausibility score selects the best hypothesis based on occupancy using a 2D height map.
Afterwards, we integrate the registration results in a continuous-time spline-based pose graph optimizer with LiDAR odometry and further sensing modalities to obtain globally consistent, georeferenced trajectories and maps. 

We evaluate the viability of our approach on multiple flights captured at two distinct testing sites.
Our method successfully reduced \ac{gnss} offset errors from up-to \SI{16}{\metre} to below \SI{0.5}{\metre} on multiple flights.
Furthermore, we obtain globally consistent maps \wrt prior 3D geospatial models.
\end{abstract}

\section{Introduction}
Georeferenced maps are essential for modern \ac{SAR} missions~\cite{kruijff2021ssrr,surmann2024jfr}.
From the initial planning stage to preserving a situational overview throughout the operation, first responders continuously compile status updates from diverse sources.
Nowadays, \acp{UAV} are well-established for aerial overviews or imaging from difficult to reach viewpoints without endangering people's life~\cite{mahmoudZadeh2024uav,beul2022fr,lyu2023uav,rosu2019ssrr,martinezAlpiste2021uav,quenzel2019jint,
hildmann2019drones,beul2018ral}.

\begin{figure}[th!]
  \centering
  \resizebox{1.0\linewidth}{!}{\input{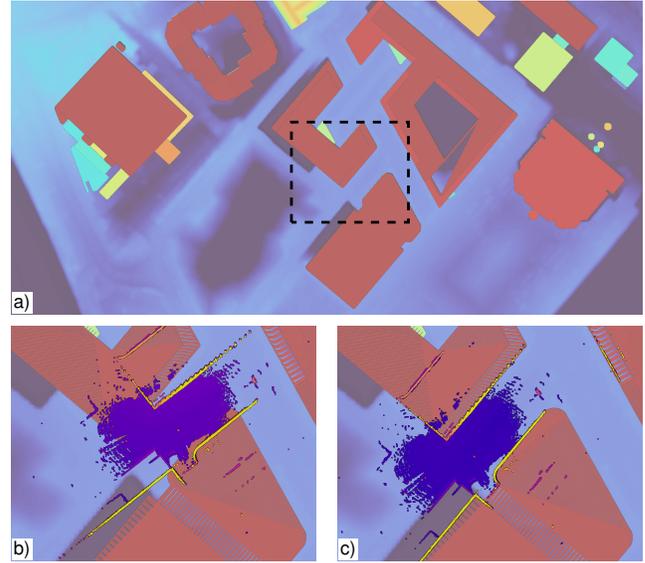}}
  \caption{Geospatial maps [a)] contain approximate building shapes and ground height. Inaccurate raw \ac{gnss} measurements impair the accuracy of georeferenced maps [b)]. We obtain a globally consistent map [c)] by registration against the geospatial model.}
  \label{fig:teaser}
\end{figure}

Automated \ac{UAV} surveys require \ac{gnss} availability not only for integration with georeferenced maps, but also for safe flight in free airspace.
Unfortunately, the precision drastically degrades in the vicinity of buildings due to reflections, shadowing and canyon effects~\citep{osm}.
This makes pure \ac{gnss}-based localization unreliable in urban areas where first responders need to fly in-between buildings.
However, LiDAR-based odometry provides locally precise poses in the proximity of 3D structures, but is prone to accumulate drift over time and lacks georeferencing.
In contrast, geospatial maps, \eg \acs{cgml} and \acs{DEM}, have accurate georeferenced poses for approximate building shapes and ground surfaces.

The registration of LiDAR maps with geospatial models promises to enhance localization in urban areas where \ac{gnss} accuracy is low and LiDAR data is informative.
Moreover, joint optimization, \eg in a pose graph, reduces drift and provides strong priors for allocentric mapping.

To enable alignment of model and LiDAR data, \acs{cgml} data is combined with the corresponding \acs{DEM}.
The LiDAR odometry~\cite{quenzel2021mars} processes scans to obtain local maps and poses.
After semantic segmentation~\cite{bultmann2023semanticfusion}, we retain only walls and ground surfaces for the georeferencing. 
\ac{gnss} and IMU measurements initialize the approximate UAV pose --- in general with an accuracy of few meters, resp. degrees.
In a next step, we perform a grid search on the coarse horizontal offset and register the local map against the model at the offsetted poses using MARS~\cite{quenzel2021mars}.
A plausibility check determines the best matching result using an intuitive score from ray-traced occupancy with a 2D height map.
The refined \ac{gnss} pose now georeferences the local LiDAR map and enables reliable registration of local maps for loop-closing.

For allocentric mapping, we directly optimize the B-spline knots of a continuous-time trajectory~\cite{sommer2020cvpr} using a pose graph.
Odometry constraints connect scans to their local map.
Additional constraints stem from relative transformation between local maps and preintegrated IMU~\cite{usenko2020basalt}.
We use the refined \ac{gnss} poses as anchors in our pose graph to obtain globally consistent trajectories and allocentric maps with accurate georeferencing.

Our approach successfully reduces the \ac{gnss} error for the whole trajectory and even for local maps and single scans.
In short, our contributions include:
\begin{itemize}
\setlength{\itemindent}{0em}
\item a refinement strategy for \ac{gnss} measurements using LiDAR registration against georeferenced 3D models,
\item an intuitive plausibility score using height-based occupancy,
\item an allocentric spline-based pose graph optimizer for continuous-time trajectories.
\end{itemize}

\section{Related Work}
In recent years, several approaches have been developed to improve the quality of \ac{gnss} data using 3D models.
\citet{cappelle2012virtual} match RGB images against textured models and fuse \ac{gnss}, odometry and gyroscope in an \ac{ukf}.
\citet{wang2019floorplan} tackle global localization on floor plans by matching corners against vertical edges within point clouds.

\citet{zhang2021seamless} consider an autonomous driving scenario and combine LiDAR point clouds with \ac{gnss}. 
Instead of registering their LiDAR measurements against externally available 3D models, they make use of maps from previous runs that have been accurately georeferenced. The registration uses a deep neural network and measurements are directly fused within an \ac{ekf}.

\citet{lucks2021improving} follow an approach similar to ours and register LiDAR scans against 3D models with the goal of mitigating the shortcomings of \ac{gnss}. As in our work, \ac{cgml} is used in combination with a \ac{DEM}. 
A major difference is their use of point-to-plane correspondences between raw scan and 3D model over long segments. Additionally, the transition between trajectory segments is interpolated to obtain a coherent map.
Instead, we use adaptive surfel maps~\cite{quenzel2021mars} and continuously register smaller segments if geometrical constraints allow successful registration.

\citet{lv2021clins} correct a continuous-time trajectory from a traditional pose graph of keyframes while maintaining the initial velocity prior to optimization.
However, we directly optimize the trajectory and further include \ac{gnss} measurements as well as relative pose between previously refined local maps.

\section{Method}
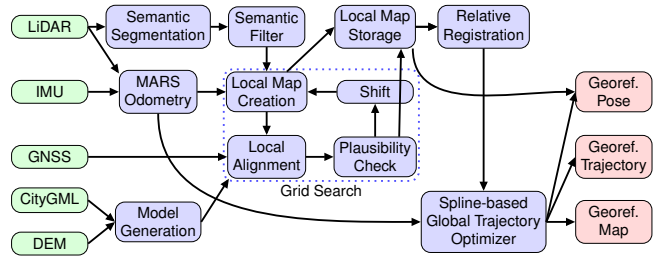
\begin{figure}
  \centering
  \resizebox{1.0\linewidth}{!}{\begin{tikzpicture}
[content_node/.append style={font=\sffamily,minimum size=1.5em,minimum width=7.1em,draw,align=center,rounded corners,scale=0.65},
vis_node/.append style={font=\sffamily,minimum size=1.5em,minimum width=5.0em,draw,align=center,rounded corners,scale=0.65},
label_node/.append style={font=\sffamily,scale=0.5},
group_node/.append style={font=\sffamily,dotted,align=center,rounded corners,inner sep=0.1em,thick,opacity=0.25},
>={Stealth[inset=0pt,length=4pt,angle'=45]},
node distance=1.5cm]

\pgfmathsetmacro{\dH}{2.00cm}

\node(LIDAR)[vis_node,fill=green!15!white] {LiDAR};
\node(IMU)[vis_node,fill=green!15!white, below of=LIDAR,yshift=0*\dH] {IMU};
\node(GNSS)[vis_node,fill=green!15!white,below of=IMU,yshift=0*\dH] {GNSS};
\node(CGML)[vis_node,fill=green!15!white,below of=GNSS,yshift=0.25*\dH] {CityGML};
\node(DEM)[vis_node,fill=green!15!white,below of=CGML,yshift=0.25*\dH] {DEM};

\node(SEM)[vis_node,fill=blue!15!white,right of=LIDAR,xshift=0.5*\dH] {Semantic\\Segmentation};
\node(MARS)[vis_node,fill=blue!15!white,right of=IMU,xshift=0.5*\dH] {MARS\\Odometry};

\node(FIL)[vis_node,fill=blue!15!white,right of=SEM,xshift=0.5*\dH] {Semantic\\Filter};
\node(LM)[vis_node,fill=blue!15!white,right of=MARS,xshift=0.5*\dH] {Local Map\\Creation};

\node(SH)[vis_node,fill=blue!15!white,right of=LM,xshift=0.5*\dH] {Shift};

\node(LA)[vis_node,fill=blue!15!white,below of=LM,xshift=0*\dH] {Local\\Alignment};

\node(PC)[vis_node,fill=blue!15!white,right of=LA,xshift=0.5*\dH] {Plausibility\\Check};

\node(AL)[vis_node,fill=blue!15!white,right of=FIL,xshift=0.5*\dH] {Local Map\\Storage};

\node(RR)[vis_node,fill=blue!15!white,right of=AL,xshift=0.5*\dH] {Relative\\Registration};

\node(PGO)[vis_node,fill=blue!15!white,below of=RR,yshift=-1.5*\dH] {Spline-based\\Global Trajectory\\Optimizer};

\node(MG)[vis_node,fill=blue!15!white,right of=CGML,xshift=0.5*\dH,yshift=-0.25*\dH] {Model\\Generation};

\node(GS)[group_node,opacity=0.8,draw=blue!85!white,fill=none,fit={(LM)(LA)(SH)(PC)}] {};
\node(GSL)[vis_node,draw=none,below of=GS,yshift=0.*\dH]{Grid Search};

\node(GM)[vis_node,fill=red!15!white,right of=PGO,xshift=0.75*\dH] {Georef.\\Map};
\node(GSP)[vis_node,fill=red!15!white,above of=GM,yshift=0*\dH] {Georef.\\Trajectory};
\node(GP)[vis_node,fill=red!15!white,above of=GSP,yshift=0.0*\dH] {Georef.\\Pose};

\draw[->,thick] (LIDAR.0) -- (SEM.180);
\draw[->,thick] (SEM.0) -- (FIL.180);
\draw[->,thick] (LIDAR.0) -- (MARS.155);
\draw[->,thick] (MARS.0) -- (LM.180);
\draw[->,thick] (FIL.270) -- (LM.90);
\draw[->,thick] (IMU.0) -- (MARS.180);
\draw[->,thick] (GNSS.0) -- (LA.180);
\draw[->,thick] (MG.0) -- (LA.210);
\draw[->,thick] (CGML.0) -- (MG.180);
\draw[->,thick] (DEM.0) -- (MG.180);
\draw[->,thick] (LM.45) -- (AL.180);
\draw[->,thick] (LA.0) -- (PC.180);
\draw[->,thick] (LM.270) -- (LA.90);
\draw[->,thick] (SH.180) -- (LM.0);
\draw[->,thick] (PC.90) -- (SH.270);
\draw[->,thick] (PC.40) -- (AL.320);
\draw[->,thick] (AL.0) -- (RR.180);
\draw[->,thick] (AL.0) -- (RR.180);
\draw[->,thick] (RR.270) -- (PGO.90);
\draw[->,thick] (MARS.270) to[out=270,in=180] (PGO.180);

\draw[->,thick] (PGO.0) -- (GP.180);
\draw[->,thick] (AL.330) to[out=270,in=180] (GP.180);
\draw[->,thick] (PGO.0) -- (GSP.180);
\draw[->,thick] (PGO.0) -- (GM.180);

\end{tikzpicture} }
  \caption{System overview: Our refinement aligns small LiDAR maps against a geospatial model using \ac{gnss} for initialization. After pose graph optimization, our system outputs a globally consistent and georeferenced map and trajectory.}
  \label{fig:pipeline}
\end{figure}

Our method consists of several steps as shown in \cref{fig:pipeline}. We describe each step in the following and start with the description of the georeferenced model.

\subsection{Georeferenced Model}
Fortunately, many German state governments make geodetic data publicly available\footnote{\url{https://www.citygmlwiki.org/index.php?title=Open_Data_Initiatives_in_Germany}}.
Our georeferenced model combines a \ac{cgml} model with a \ac{DEM}. The state of North Rhine-Westphalia publishes both in \SI{1}{\kilo\metre\squared} sized tiles\footnote{\url{https://www.opengeodata.nrw.de/produkte/geobasis/}}.
The \ac{cgml} contains the rough shape of buildings with \ac{lod}-2 whereas the \ac{DEM} contains the grounds' height in a grid with a \SI{1.0}{\metre\squared} resolution.
At first, we convert the \ac{cgml} model into a triangular mesh\footnote{\url{https://github.com/citygml4j/citygml-tools}} and subdivide the triangles~\cite{seeger2001triangle} until their area is at most \SI{0.1}{\metre\squared}. Afterwards, we retain the triangle vertices.
Similarly, we bi-linearly interpolate the \ac{DEM} grid to \SI{0.1}{\metre\squared} resolution and merge it with the sampled \ac{cgml} points for further processing.

\begin{figure*}
\centering
{\input{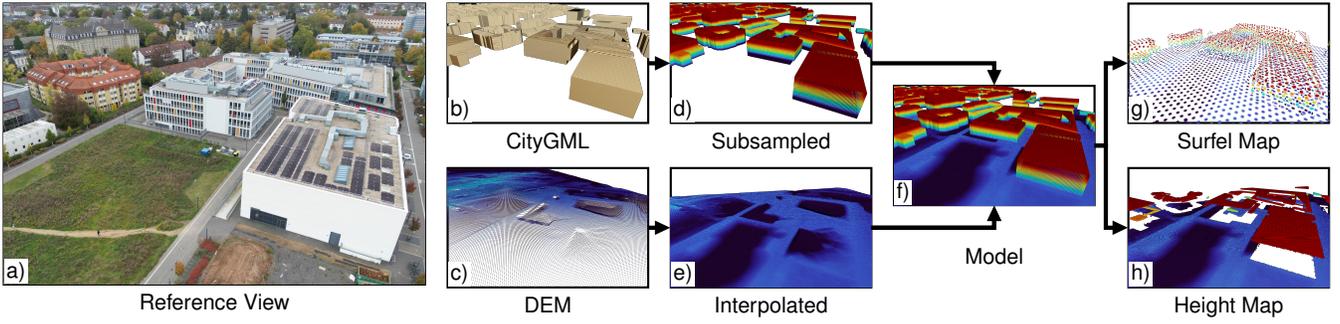}}
\caption{Model Generation: A view [a)] on the Poppelsdorf campus at the University of Bonn for easier scene understanding. The \acs{cgml} data [b)] contains the rough polygonal building shape, while the \ac{DEM} [c)] represents the ground surfaces. We combine the subsampled polygonal \ac{cgml} [d)] with the interpolated \ac{DEM} [e)] for our model [f)]. A surfel map [g)] is derived for registration~\cite{quenzel2021mars} and a height map [h)] for our plausibility check.}
\label{fig:modelgen}
\end{figure*}

If the \ac{cgml} model is not available or the model quality appears insufficient after visual inspection, we extract ``roof'' and cluster ``contour'' annotated points from \ac{ALS}.
After identifying ``roof'' points close to the contour, we extract the roof's 2D $\alpha$-shape without height using CGAL~\cite{tran2024cgal} and sample points in a line vertically from the roof down to the floor height.

For registration, MARS~\cite{quenzel2021mars} derives a multi-resolution surfel map from the point cloud.
Additionally, we compute a 2D height map where each cell stores the maximum height.
This height map aids to assess the quality of the registration result with our plausibility score~(\cref{subsec:check}). 
\cref{fig:modelgen} depicts the model of the Poppelsdorf Campus at the University of Bonn.

\subsection{Scan Preparation}
\label{subsec:dataprep}
\begin{figure}
  \centering
  \resizebox{1.0\linewidth}{!}{\input{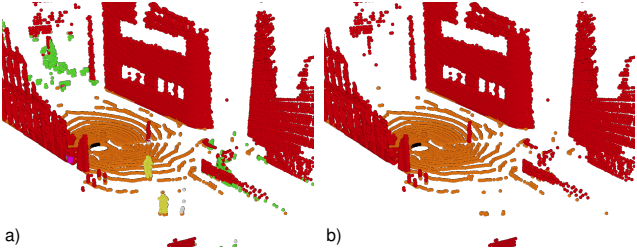}}
  \caption{A semantically annotated LiDAR scan with vegetation (green) and people (yellow) before [a)] and after filtering [b)].}
  \label{fig:segmentation}
\end{figure}

Our georeferenced model only contains ground and building surfaces.
However, moving people, vegetation or other obstacles may be present in the actual LiDAR scans.
Hence, we filter out clutter and retain only ground and building points using semantic segmentation~\cite{bultmann2023semanticfusion}, as shown in \cref{fig:segmentation}.

Filtered single scans $\mathcal{P}$ are very sparse and more difficult to register. Thus, aggregating multiple scans into local maps $\mathcal{W}$ creates more complete surfel maps. As measurements from the same position are redundant, the \ac{UAV} needs to move more than $\tau$ \si{\metre} since the last accumulated scan.

\subsection{Georeferenced Local Alignment}
\label{subsec:alignment}
Given a local map and initial pose, we align its surfel map against the model's surfel map with the registration of MARS~\cite{quenzel2021mars} optimizing a single pose.
Our initial guess for the local map pose stems from data provided by the \ac{UAV}. 
A 3-axis magnetometer inside the IMU provides the orientation, whereas the \ac{gnss} supplies the approximate horizontal position and optionally the altitude. 
In proximity to the ground, we initialize the height from the model's height map and the ultrasonic sensor of the \ac{UAV}.

During tests, the horizontal \ac{gnss} position sometimes differed from the actual position by as much as \SI{16}{\metre}.
Such error exceeds the convergence radius for local registration and results in convergence to non-global local minima.
We mitigate this with a grid search on the coarse horizontal shift and run the registration for each shifted initial pose.
The grid covers the uncertainty range, \eg with a resolution of \SI{2}{\metre}, such that the local convergence basins overlap.
Afterwards, a subsequent plausibility check (\cref{subsec:check}) determines the best local minima.

In the absence of a magnetometer, the IMU accelerometer allows computation of roll and pitch whereas the grid search extends to combinations of horizontal shift and yaw.

\subsection{Plausibility Check}\label{subsec:check}
Each alignment from the aforementioned grid search has to be checked for its plausibility to reject incorrect local minima.
We propose a ray tracing-based plausibility score that operates solely on a discretized height map calculated from the model.
The basic idea is to compare measured LiDAR rays with the corresponding projected rays in the model at the aligned pose.

After voxel filtering the scan points, we ray-trace horizontally using Bresenham's line algorithm~\cite{bresenham1965algorithm} from the sensor position $\bm{o}$ towards the point $\bm{p}$ in the height map. 
For every cell along the ray, we check that the ray $h_{\bm{p}_r}$ is above the height map $h_{\bm{p}_m}$.
With the discretized ray distance $d_{\bm{p}}$ from $\bm{o}$ to $\bm{p}$ and the model distance $d_{\bm{p}_m}$ from $\bm{o}$ to the first intersection ($h_{\bm{p}_m} > h_{\bm{p}_r}$), we compute a ray-score $c_{\mathrm{ray}}$ as follows:
\begin{align}
c_\mathrm{ray}(\bm{p}) &= \min\left(\frac{d_{\bm{p}_m}}{d_{\bm{p}}}, 1\right) \in \left[0,1\right].
\end{align}
The score $s_\mathrm{ray}$ increases linearly with the measured ray distance up until the length of the projected ray in the model.
Here, an upper limit of \num{1} ensures that measuring further, \eg through windows, is neither penalized nor encouraged as the \ac{lod}-2 model only contains the rough shape of the facade and not its interior.

On its own, this would lead to incorrect results when comparing measurements on an open field with a building in the model. 
Hence, we introduce a binary hit score $c_{\mathrm{hit}}$:
\begin{align}
 c_\mathrm{hit}(\bm{p}) &= \begin{cases}
   1, & \text{if } \left(h_{\bm{p}_m} > \left(h_{\bm{p}} + \varepsilon\right)\right) \land \left( \lvert d_{\bm{p}_m} - d_{\bm{p}}\rvert < \vartheta \right),\\
   0, & \text{otherwise}.
 \end{cases}
\end{align}
Intuitively, it is implausible to measure an intersection ($h_{\bm{p}_m} > h_{\bm{p}}$) if there is no obstacle in the model map. At the same time, 
the endpoint should be close to the surface ($\lvert d_{\bm{p}_m} - d_{\bm{p}}\rvert < \vartheta$).

We obtain our plausibility score $s_\mathcal{W}$ for a local map as the mean over all scores with a linear combination of $c_\mathrm{ray}$ and $c_\mathrm{hit}$ with weight $w\in[0,1]$:
\begin{align}
c_{\bm{p}} &= w\, c_{\mathrm{ray}}(\bm{p}) + (1-w)\, c_{\mathrm{hit}}(\bm{p}),\\
s_\mathcal{W} &= \frac{1}{\abs{\mathcal{W}}} \sum_{\mathcal{P}\in\mathcal{W}} \frac{1}{\abs{\mathcal{P}}} \sum_{\bm{p}\in\mathcal{P}} c_{\bm{p}}.
\end{align}
As apparent from construction, the most plausible alignment from~\cref{subsec:alignment} should have $s_\mathcal{W}$ closest to one, which necessitates both criteria being close to one for all points.

\begin{figure*}[ht!]
  \centering
  \resizebox{1.0\linewidth}{!}{\input{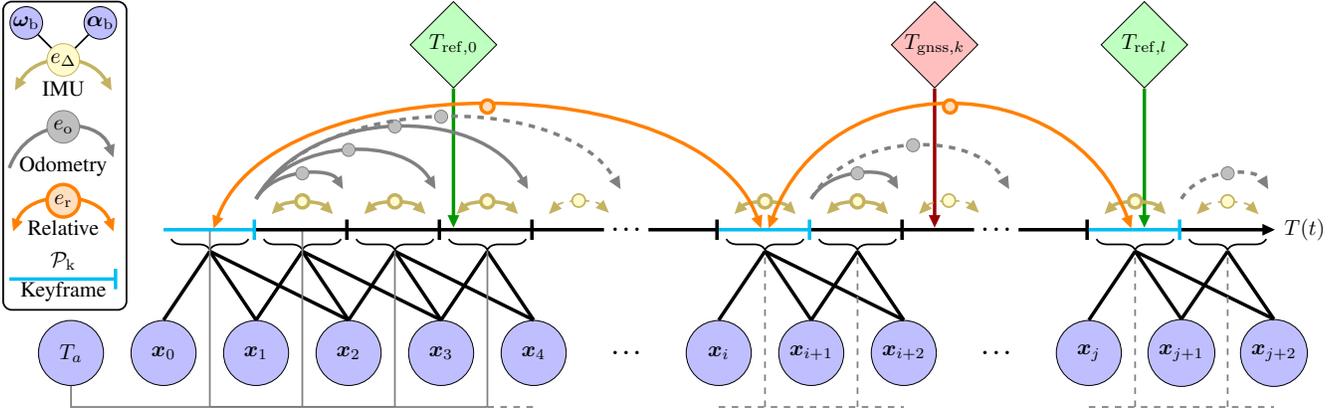}}
  \caption{Spline-based Pose Graph:
We estimate a continuous-time B-spline trajectory~\cite{sommer2020cvpr} $T(t)$ with $N$ knots ($\bm{x}_i,\ldots,\bm{x}_{i+N-1} \in \mathcal{X}$) being active per scan.
Raw and refined \ac{gnss} positions ($T_\mathrm{gnss},T_\mathrm{ref}$) allow to georeference the \ac{UAV} trajectory with an anchor pose $T_\mathrm{a}$.
Odometry constraints ($e_\mathrm{o}$) connect each scan $\mathcal{P}$ with the previous keyframe.
Preintegrated IMU measurements ($e_\mathrm{\Delta}$) with optimizable biases ($\bm{\omega}_\mathrm{b},\bm{\alpha}_\mathrm{b}$) enforce smoothness within a scan.
Relative poses ($e_\mathrm{r}$) between keyframes or scans enable loop closing.}
  \label{fig:graph}
\end{figure*}

Additionally, we compute the covariance's condition numbers from model surfel map to local surfel map and vice versa to detect possible slippage of the translation components during registration. Testing both directions ensures rejection even if associations differ.
If $\kappa$ is below a threshold $\tau$ and $s_\mathcal{W}$ is above $\gamma$, we accept the refined \ac{gnss} pose $T_\mathrm{ref}$ for the local map.

\subsection{Spline-based Global Trajectory Optimization}

In order to obtain a globally consistent map and trajectory, we build a pose graph $(\mathcal{V},\mathcal{E})$ that represents the full \ac{UAV} path using a continuous-time B-spline $T_\mathcal{X}(t)$ as in MARS~\cite{quenzel2021mars}.
In contrast to CLINS~\cite{lv2021clins}, we do not build a standard pose graph from some keyframes and optimize the keyframe poses only. 
Instead, our graph vertices $\mathcal{V}$ directly contain the B-spline knots $\mathcal{X}$, the \ac{gnss} anchor pose $T_\mathrm{a}$, IMU accelerometer and gyroscope biases ($\bm{\alpha}_\mathrm{b},\bm{\omega}_\mathrm{b}$), as shown in \cref{fig:graph}.
The allocentric pose $T(t)$ at time $t$ is given by:
\begin{align}
T(t) &= T_\mathrm{a}  T_\mathcal{X}(t).
\end{align}

We jointly minimize for all constraints $e\in \mathcal{E}$ the Mahalanobis error with distance $\bm{d}_e$ and covariance $\Sigma_e$ using a robust huber norm~\cite{huber1964robust}:
\begin{align}
\argmin_\mathcal{V} \sum_{e\in \mathcal{E}} \rho_\mathrm{huber} \left( \bm{d}_e^\intercal{\Sigma_e^{-1}}\bm{d}_e \right).
\end{align}

A raw or refined \ac{gnss} pose $T_\mathrm{abs}$ with covariance $\Sigma_\mathrm{abs}$ provides an absolute constraint $e_\mathrm{a} \in \mathcal{E}$ on $T(t)$:
\begin{align}
\bm{d}_\mathrm{a} &= Log_{SE(3)}\left(T_\mathrm{abs}^{-1} T_\mathrm{a} T_\mathcal{X}(t)\right) \in \mathbb{R}^6,
\end{align}
using the logarithm map $Log_{SE(3)}$~\cite{sommer2020cvpr} .
Alternatively, this may be restricted to only the position $\bm{p}_{\mathrm{abs}}$:
\begin{align}
\bm{d}_{\mathrm{a},\bm{p}} &= \left(T_\mathrm{a} \bm{p}_\mathcal{X}(t) - \bm{p}_\mathrm{abs}\right)\in\mathbb{R}^3.
\end{align}

Odometry constraints $e_o\in \mathcal{E}$ connect from scan at time $t_s$ towards the previous keyframe at $t_k$ with pose $T_\mathrm{o}$: 
\begin{align}
\bm{d}_\mathrm{o} &= Log_{SE(3)}\left(T_\mathrm{o}^{-1} T_\mathcal{X}(t_k)^{-1} T_\mathcal{X}(t_s)\right) \in \mathbb{R}^6.
\end{align}
Preintegrated IMU measurements~\cite{usenko2020basalt} $e_{\Delta} \in \mathcal{E}$ connect consecutive scans from $T_\mathcal{X}(t_{s-1})$ to $T_\mathcal{X}(t_{s})$.

Additional relative pose constraints $e_r \in\mathcal{E}$ with $T_\mathcal{X}(t_1) \approx T_\mathcal{X}(t_0) T_\mathrm{rel}$ stem from registration of time-wise or spatially neighboring local maps:
\begin{align}
\bm{d}_r &= Log_{SE(3)}\left(T_\mathrm{rel}^{-1} T_\mathcal{X}(t_0)^{-1} T_\mathcal{X}(t_1)\right) \in\mathbb{R}^6.
\end{align}
Here, our refined \ac{gnss} poses $T_{\mathrm{ref}}$ aid in identifying spatially neighboring maps and initialize the relative pose $T_\mathrm{rel}$ for registration.
We only add relative constraints if the initial translational distance for $\bm{d}_r$ is smaller than \SI{5}{\percent} of the distance along the trajectory.
This intuitively allows larger deviations for more distant loop-closures as errors accumulate over time.

Prior to optimization, we initialize yaw and horizontal 2D position of the \ac{gnss} anchor pose $T_\mathrm{a}$ by aligning~\cite{eggert1997align} the refined \ac{gnss} positions with the corresponding spline positions.
Empirically, we found this to provide a better initialization over longer segments than using a single pose since the IMU orientation might be slightly incorrect.

\begin{figure*}[ht!]
\centering
  \resizebox{1.0\linewidth}{!}{\input{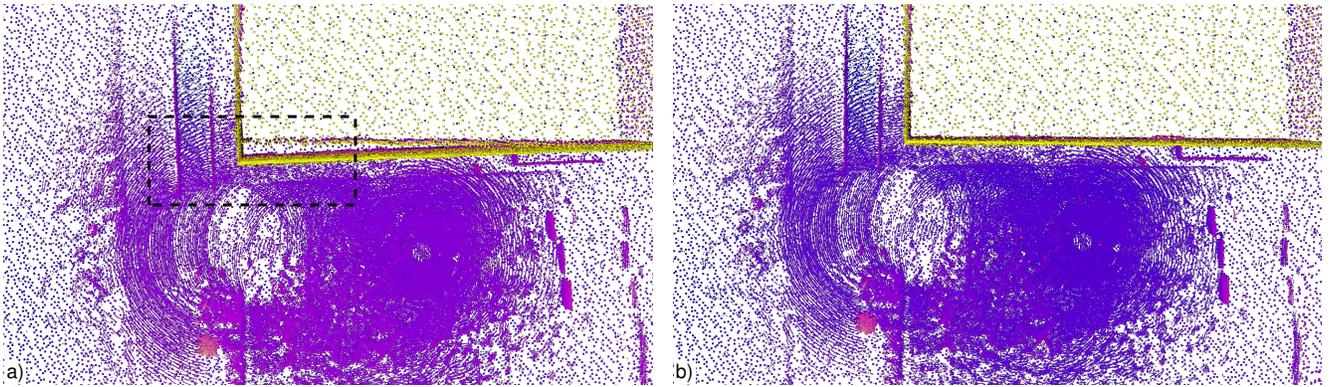}}
\caption{Top-down closeup after optimization using only raw [a)] or refined \ac{gnss} [b)]. Our refinement successfully corrected the \ac{gnss} offset and reduced the gap between height-colored model and local map.}
\label{fig:closeup}
\end{figure*}

\section{Evaluation}
We recorded multiple \ac{UAV} flights on different days at the Campus Poppelsdorf of the University of Bonn.
Our \ac{UAV}~\cite{schleich2021drz} is a modified DJI M210v2 equipped with an Intel NUC and an Ouster OS-0 128 LiDAR. The \ac{UAV} has an external \ac{gnss} antenna by DJI to increase separation from the compute hardware and reduce possible interference.

Although, one can expect higher accuracy from RTK- and D-\ac{gnss}, our \ac{UAV} is not equipped with either.
Nonetheless, we recently experienced occasional offsets above \SI{1}{\metre} and up to \SI{4}{\metre} during static positioning tests with a Holybro UM982 RTK-GNSS due to canyoning.

We align the final maps against a georeferenced \ac{TLS} cloud of the campus using CloudCompare\footnote{\url{https://cloudcompare.org/}}. Afterwards, we compute the RMS positional error for the raw \ac{gnss} and refined \ac{gnss} measurements. Additionally, we evaluate the RMS positional error after optimization evaluated at the LiDAR scan time. The results are shown in \cref{tab:results}.

\begin{figure*}[ht!]
\centering
\resizebox{1.0\linewidth}{!}{\input{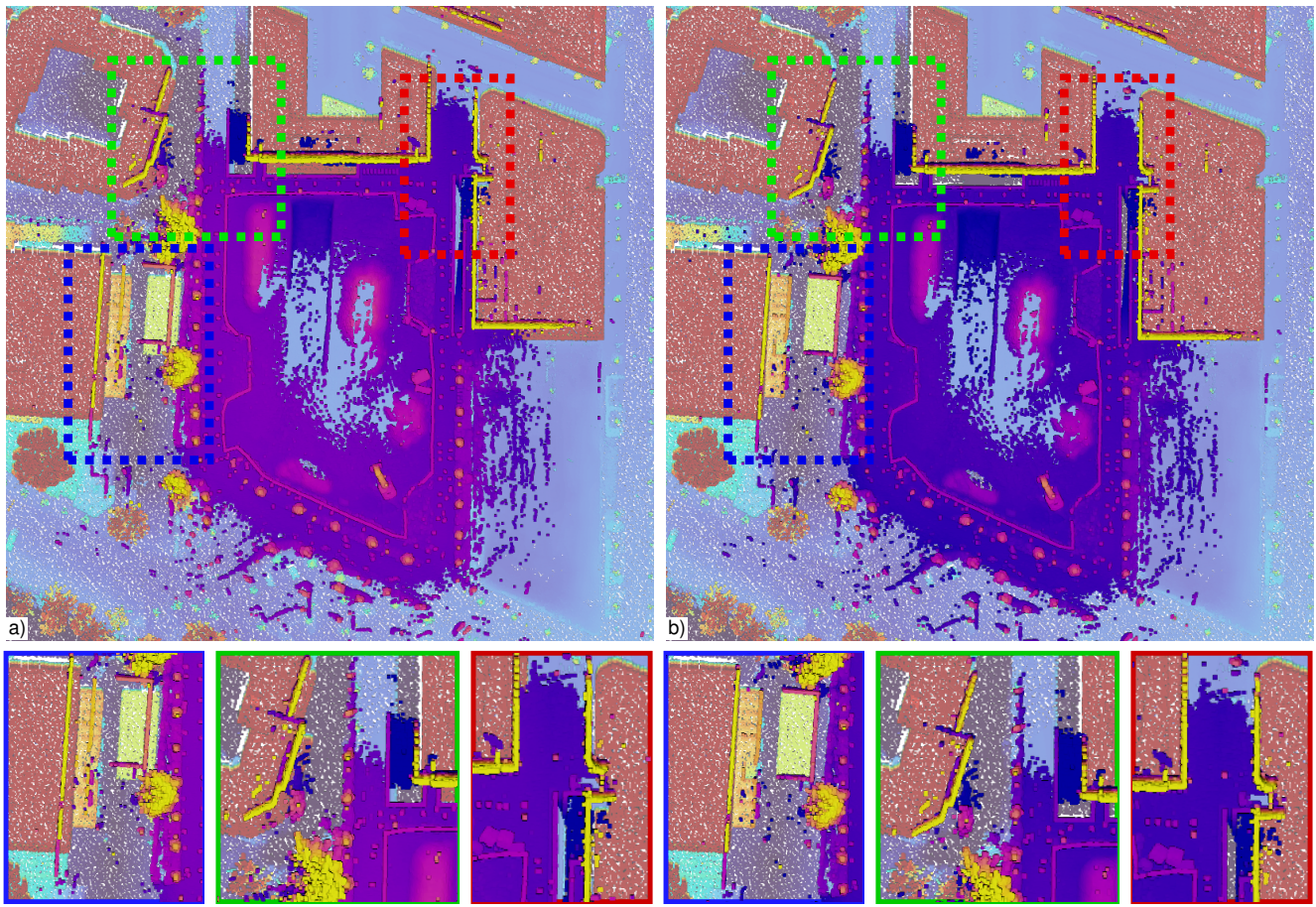}}
\caption{Top-down view of the Poppelsdorf Campus at the University of Bonn after optimization using only raw [a)] or refined \ac{gnss} [b)]. For raw \ac{gnss}, facades (\eg yellow lines) are mapped inside buildings (\eg red areas). Our optimization successfully corrected the \ac{gnss} offsets and obtained a globally consistent map showing aligned facades \wrt the geospatial height-colored \ac{ALS} map.}
\label{fig:overview}
\end{figure*}

\cref{fig:closeup} highlights the difference between using raw and refined \ac{gnss} for georeferencing the reconstructed point cloud. The visible gap for raw \ac{gnss} vanishes after optimization when using refined \ac{gnss}.
\cref{fig:overview} further emphasizes this on sequence ``16-38-18'' for a larger section of the campus, showing clear improvement in aligning building walls.

The grid search for all offsets including alignment takes around \SI{0,5}{\second} using a radius of \SI{8}{\metre} and \SI{4}{\metre} step size. Hence, the search can run in parallel during normal operation since new local maps are only created sporadicly.
Our spline-based optimizer takes around \SI{1.5}{\second} for the \SI{166}{\second} long sequence ``11-17-39'' (\cref{fig:closeup}).

\bgroup
\renewcommand{\arraystretch}{1.2}  
\begin{table}
\small
\centering
\caption{RMSE Evaluation \wrt georeferenced TLS}
\begin{tabularx}{\columnwidth}{K{2.5cm}ZZZZ}
\toprule
Sequence &  raw GNSS [\si{\metre}] & ref. GNSS [\si{\metre}] & opt. w/ raw GNSS [\si{\metre}] & opt. w/ ref. GNSS [\si{\metre}]\\
\midrule\midrule
11-13-36 & 1.896 & 0.039 & 1.808 & 0.033 \\\hdashline
11-17-39 & 0.813 & 0.575${}^{\ast}$ & 0.777 & 0.053 \\\hdashline
14-18-47 & 461.861${}^{\dagger}$ & 0.295 & 443.965 & 0.062 \\\hdashline
16-38-18 & 2.081 & 0.091 & 2.082 & 0.078 \\\hdashline
16-44-52 & 4.296 & 0.145 & 3.618 & 0.106 \\\hdashline
17-24-39 & 17.664 & 0.150 & 1.701 & 0.055\\
\bottomrule
\end{tabularx}
\vspace{0.25cm}${}^\ast$ due to outlier, w/o: 0.057 \si{\metre}. ${}^\dagger$ contains mostly outlier.
\label{tab:results}
\end{table}
\egroup

\begin{figure*}[ht!]
\centering
\resizebox{1.0\linewidth}{!}{\input{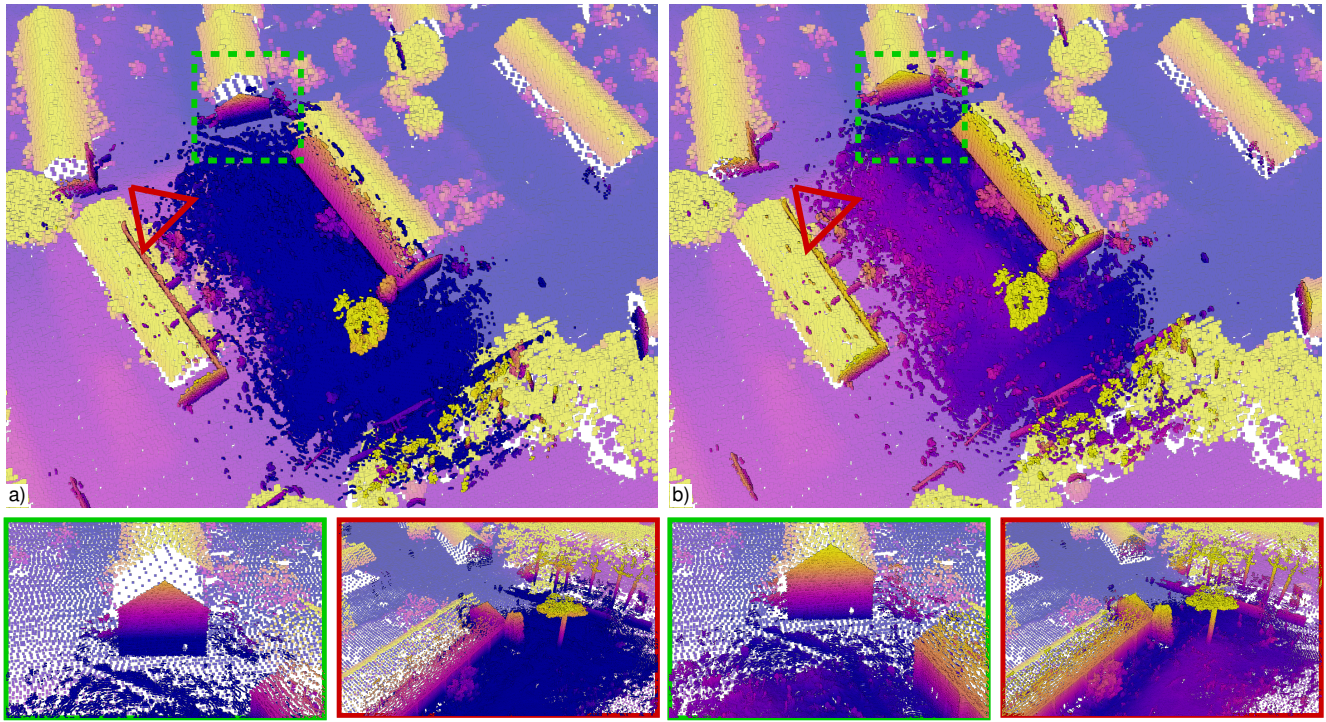}}
\caption{Georeferenced maps after optimization using only raw [a)] or refined GNSS [b)] are overlaid on \ac{ALS} reference points and color-coded by height. Sequence captured during a forest fire training exercise by the fire brigade of the district Viersen at the abandoned Javelin Baracks in Elmpt, Germany. \ac{ALS} points are enlarged for better visibility.
The gap between roof and measured wall (green rectangle) vanishes after optimization with refined GNSS.
Large portions of the scene, as viewed from the red triangle, consist of vegetation which is non-represented in the \ac{cgml} and \ac{DEM}.}
\label{fig:viersen}
\end{figure*}

We recorded 3 additional sequences at the abandoned Javelin Baracks in Elmpt, Germany during a forest fire training exercise by the fire brigade of the district Viersen.
The \ac{ALS} predominantly contains ground, plant, tree top and roof measurements with hardly any walls at a point density of $\approx$ \num{4} to \num{10} points per \si{\metre\squared} and an accuracy below \SI{30}{cm}.
In contrast to the previously used \ac{TLS}, the \ac{ALS} cloud exhibits too little overlap with our measurements to reliably constrain a reference alignment in lateral direction with sufficient accuracy.
As a result, we showcase our results in \cref{fig:viersen} and report the positional RMS distance for raw and refined \ac{gnss} \wrt our estimated correction.
On average, our pipeline corrected the raw \ac{gnss} between \SI{2.40}{\metre} and \SI{3.06}{\metre} per sequence, whereas most refined \ac{gnss} positions are off by less than \SI{10}{\centi\metre}.

\section{Conclusion}
We presented a novel approach to register local LiDAR maps against geospatial data to reduce \ac{gnss} offsets.
A ray-tracing based score allows to select plausible refined \ac{gnss} poses.
Our new spline-based global trajectory optimizer delivers globally consistent allocentric 3D maps.
Our experiments showcased the effectiveness of our approach and successfully reduced the \ac{gnss} offset from multiple meters to below \SI{0.5}{\metre}.

\section*{Acknowledgement}
We would like to thank Martin Blome and Lasse Klingbeil from the Institute of Geodesy and Geoinformation of the University of Bonn for the \acs{TLS} reference map. Moreover, we would like to express our gratitude to the fire brigade of the district Viersen for the oportunity to participate in their forest fire training exercises.
This work has been supported by the German Federal Ministry of Education and Research (BMBF) in the projects ``Kompetenzzentrum: Etablierung des Deutschen Rettungsrobotik-Zentrums (E-DRZ)'', grant 13N16477, and ``UMDenken: Supportive monitoring of turntable ladder operations for firefighting using IR images'', grant 13N16811.

\printbibliography
\balance

\end{document}